\documentclass[a4paper]{svproc}
\usepackage{url}
\usepackage{amsmath}
\usepackage{amsfonts}
\usepackage{graphicx}
\usepackage{placeins}

\usepackage[dvipsnames]{xcolor} 

\usepackage{tikz}
\usepackage{amsmath}
\usetikzlibrary{arrows.meta, shapes.geometric, positioning, fit, backgrounds, calc}
\usepackage{graphicx}

\begin{document}
\mainmatter              
\title{Hierarchical Admittance-Impedance Control for Mobile Manipulator Load Transfer}
\titlerunning{Hierarchical Admittance–Impedance Control}  
%
\author{Xu Minghao\inst{1} \and Thushara Sandakalum\inst{2} \and \\ Christina Dao Wen Lee\inst{2} \and Marcelo H. Ang Jr.\inst{2}}
\authorrunning{Xu Minghao et al.} 
%
%
\institute{NUS Graduate School,\\
National University of Singapore, Singapore 119077,\\
\email{xu\_minghao@u.nus.edu}
\and
Department of Mechanical Engineering,\\
National University of Singapore, Singapore 119077}

\maketitle              

\begin{abstract}

This paper presents a hierarchical admittance--impedance controller that combines an inner Cartesian impedance controller, an outer 6D admittance supervisor, and workspace-aware mobile-base assistance for guided load transfer. The method is motivated by assistive tasks such as airport baggage handling, where a robot should support a payload while a human worker guides its motion through small physical forces. Following an interaction-role interpretation of admittance and impedance, the torque-controlled Franka Research 3 manipulator behaves as a Cartesian impedance at the robot--object interface, providing compliant load support around a commanded equilibrium pose. At the human-facing interaction level, the outer admittance supervisor converts measured human wrench input into smooth Cartesian reference motion. The workspace-aware assistance layer shares the admittance-generated translational motion between the manipulator and a velocity-controlled base when larger motion is required, while front-side load alignment keeps the carried object near the robot's frontal region for safer and more intuitive guidance. The system was implemented in ROS and validated through a suitcase transfer experiment, showing that the robot can support the load while allowing the human to guide the transfer with relatively little effort.

\end{abstract}
\section{Introduction}

Physical human-robot collaboration requires robots to respond safely and intuitively to human-applied forces while maintaining task performance~\cite{Villani2018SurveyHRC,Haddadin2017RobotCollisions}. One motivating application is airport baggage handling, where workers repeatedly lift, pull, reposition, and transfer baggage between carts, conveyors, and aircraft-side equipment. These operations are physically demanding and often involve heavy and/or irregularly shaped objects. A mobile manipulator can assist by supporting most of the payload weight while allowing a human worker to guide the baggage through small corrective forces.

In this setting, the robot should not simply execute a fully autonomous trajectory. Instead, the robot should behave as a compliant partner that reduces human effort while preserving the worker's authority over the robot motion. The human should be able to push, pull and correct the object's pose through direct physical interaction, while the robot provides load support and whole-body motion assistance. This requires a control architecture that presents admittance-like behavior to the human while maintaining stable impedance behavior at the manipulator.

The motivation of this work follows the interaction-theoretic view of impedance and admittance introduced by Hogan~\cite{Hogan1985PartI,Hogan1985PartII,Hogan1985PartIII}. Rather than treating impedance and admittance as two alternative controller choices, this work assigns them to different interaction roles. At the robot-object interface, the manipulator behaves as an impedance, regulating the relation between displacement, velocity, and interaction wrench. At the human-robot/object interface, the combined robot-object system behaves as an admittance, where measured human force generates motion. This role assignment motivates a hierarchical architecture in which an inner impedance-controlled manipulator is guided by an outer admittance supervisor.

Prior work has established impedance control for compliant manipulation, admittance control for force-guided motion, and whole-body compliance for mobile manipulators. However, less attention has been given to a simple and implementable hierarchy that combines these ideas for human-guided load transfer, while also extending workspace through the mobile base and maintaining a favorable carrying pose.

This paper presents an implemented hierarchical admittance-impedance controller for a mobile manipulator. The platform consists of a torque-controlled Franka Research 3 (FR3) manipulator mounted on mobile base. The manipulator is governed by an inner Cartesian impedance controller, while an outer admittance supervisor converts measured interaction wrench into desired compliant end-effector motion. A workspace-aware base assistance layer then determines how much of the translational motion should be realized by the mobile base and how much should remain as a Cartesian reference update for the manipulator.

For guided load transfer, safety depends not only on compliant interaction but also on maintaining a favorable loaded configuration. Excessive manipulator extension can move the manipulator toward its workspace boundary, while lateral payload drift can create side loading and an unfavorable load distribution. The proposed supervisor, therefore, includes workspace-aware base assistance to limit manipulator reach and frontal load alignment to keep the payload near the robot's frontal support region during carrying.

This paper makes the following contributions:
\begin{enumerate}
    \item \textbf{A hierarchical admittance-impedance control framework} for human-guided load transfer with a mobile manipulator.
    
    \item \textbf{A torque-level Cartesian impedance controller} for the FR3 manipulator, that provides local compliant load support around a moving equilibrium pose.
    
    \item \textbf{A 6D outer admittance supervisor} that converts measured human interaction wrench into smooth Cartesian reference motion.
    
    \item \textbf{A workspace-aware mobile-base assistance strategy} that shares translational motion between the manipulator and the base, with frontal load alignment to keep the payload in front of the robot.
    
    \item \textbf{An experimental suitcase-transfer validation} showing that the robot can support the payload while the human provides only small guidance forces.
\end{enumerate}

\section{Related Work}

Physical human--robot interaction has long been studied through the relationship between force and motion. Hogan's impedance control formulation established manipulation as the control of dynamic interaction rather than motion or force alone~\cite{Hogan1985PartI,Hogan1985PartII,Hogan1985PartIII}. This viewpoint remains foundational for contact-rich robot control and motivates the use of mechanical behavior as a design objective in physical interaction.

Impedance control is widely used for robot-side physical interaction, especially for manipulators with torque control. Operational-space and Cartesian impedance formulations provide the foundation for regulating compliant behavior in task space~\cite{Khatib1987OperationalSpace,Ott2008CartesianImpedance,AlbuSchaeffer2003CartesianImpedance}. Cartesian impedance control establishes a physically interpretable relationship between end-effector displacement and restoring wrench, enabling compliant responses around a commanded equilibrium pose.

Admittance control converts measured external wrench into robot motion through a virtual dynamical model. Keemink et al.~\cite{Keemink2018Admittance} reviewed admittance control for physical human--robot interaction, including the effects of force filtering, motion-loop dynamics, and virtual damping on interaction behavior. This framework supports force-guided motion by modifying the reference supplied to a lower-level controller.

Adaptive and variable compliance methods have also been investigated to improve physical interaction quality~\cite{Ficuciello2015VariableImpedance}. More recently, Chen et al.~\cite{Chen2024ComplianceWhileResisting} proposed a nonlinear admittance controller inspired by shear-thickening fluid behavior and evaluated compliant guidance and resistance to impact disturbances on both fixed and mobile manipulators. Such developments highlight the importance of shaping the robot's interaction dynamics for different guidance and disturbance conditions.

Mobile manipulators extend the workspace of fixed-base manipulators by coordinating locomotion and manipulation~\cite{Yamamoto1994MobileManipulator,Seraji1998MobileManipulators}. Dietrich et al.~\cite{Dietrich2016WholeBodyImpedance} proposed whole-body impedance control for wheeled mobile manipulators, including an admittance interface for the motion-controlled base. Leboutet et al.~\cite{Leboutet2019WholeBodyCompliance} investigated tactile-based whole-body compliance using force propagation. These approaches establish the mobile base as part of the robot's compliant behavior and support physical interaction beyond the workspace of the arm alone.

More directly related to collaborative load transfer, Benzi et al.~\cite{Benzi2022PassiveTransportation} developed a whole-body control architecture for a mobile manipulator assisting a human during transportation. Their method combines variable admittance with passivity constraints and coordinates the arm and base through a whole-body kinematic model. Both subsystems receive velocity commands. This provides a relevant foundation for collaborative transportation, while the present system retains torque-level Cartesian impedance on the manipulator and coordinates it with a velocity-controlled base through supervisory reference generation.

Recent work has also extended compliant whole-body coordination to legged mobile manipulators. Schperberg et al.~\cite{Schperberg2026SafeLocoManipulation} combined model-based admittance control with a learned locomotion policy, coordinating arm and base motion to realize a six-dimensional response to external wrench input. This demonstrates the continued relevance of integrating interaction-driven reference generation with whole-body motion coordination across different mobile platforms.

Building on these developments, the present work focuses on an implementable hierarchy for human-guided load transfer. A torque-controlled manipulator provides local Cartesian impedance, an outer 6D admittance supervisor generates human-guided reference motion, and a velocity-controlled mobile base assists through workspace-aware activation and frontal load alignment. The latter functions address the task-level objectives of extending the transfer workspace and keeping the payload in a predictable carrying region. The contribution lies in integrating these elements into a complete load-transfer procedure and demonstrating their coordinated operation in a suitcase-transfer task motivated by assistive baggage handling.

\section{Problem Formulation}

\subsection{Motivating Use Case: Human-Guided Load Transfer}

The task to be completed is human-guided load transfer, inspired by airport baggage handling. In this task, the robot supports a payload while the human physically guides it from a start region to a goal region. The human is not required to carry the full payload or precisely command a trajectory. Instead, the human should be able to push, pull, or adjust the payload with small physical forces, while the robot provides compliant support to compensate for the load’s weight.


\subsection{System Description and Notations}

The mobile manipulator system consists of a Franka Research 3 manipulator mounted on a mobile base. The controller uses explicit reference frame notation. Let $\mathcal{F}_{W}$ denote the world frame, $\mathcal{F}_{B}$ the mobile base frame, $\mathcal{F}_{0}$ the FR3 link-0 frame, and $\mathcal{F}_{E}$ the end-effector frame. 

A left superscript indicates the frame in which a quantity is expressed. For example, ${}^{B}\mathbf{T}_{E}$ denotes the end effector pose expressed in $\mathcal{F}_{B}$, ${}^{0}\mathbf{V}_{E}$ denotes the end-effector twist expressed in $\mathcal{F}_{0}$, and ${}^{0}\mathbf{W}_{h}$ denotes the human interaction wrench expressed in $\mathcal{F}_{0}$.

Unless otherwise stated, all Cartesian quantities used by the controller, including poses, twists, wrenches, and reference errors, are expressed in the FR3 link-0 frame $\mathcal{F}_{0}$. This convention is consistent with the current implementation where the measured wrench is transformed into $\mathcal{F}_{0}$ before being used by the admittance supervisor.

The end-effector pose can be written as
\begin{equation}
{}^{W}\mathbf{T}_{E}
=
{}^{W}\mathbf{T}_{B}
{}^{B}\mathbf{T}_{0}
{}^{0}\mathbf{T}_{E}(\mathbf{q}),
\end{equation}
where ${}^{W}\mathbf{T}_{B}$ is the mobile base pose, ${}^{B}\mathbf{T}_{0}$ is the fixed transform from the mobile base to the FR3 link-0 frame, and ${}^{0}\mathbf{T}_{E}(\mathbf{q})$ is the FR3 forward kinematics.

\subsection{Controller Objectives}

The overall control objective is to realize intuitive human-guided load transfer with a mobile manipulator. The controller should satisfy the following requirements:
\begin{itemize}
    \item The robot should support the payload while allowing the human to guide the motion with relatively small physical effort.
    \item The force and torque applied by human should generate smooth translational and rotational motion.
    \item The FR3 manipulator should remain locally Cartesian impedance controlled.
    \item The mobile base should extend the effective transfer workspace while keeping the payload in the frontal region of the robot.
    \item The generated Cartesian reference should remain bounded and smooth during persistent force input, intermittent guidance, grasping, lifting, carrying, and release.
\end{itemize}

\section{Hierarchical Admittance-Impedance Control Architecture}

The FR3 manipulator is torque-controlled through an effort interface and runs a Cartesian impedance controller. The mobile base behavior is governed by a velocity controller.

\begin{figure}[!htbp]
    \centering
    \includegraphics[width=\linewidth]{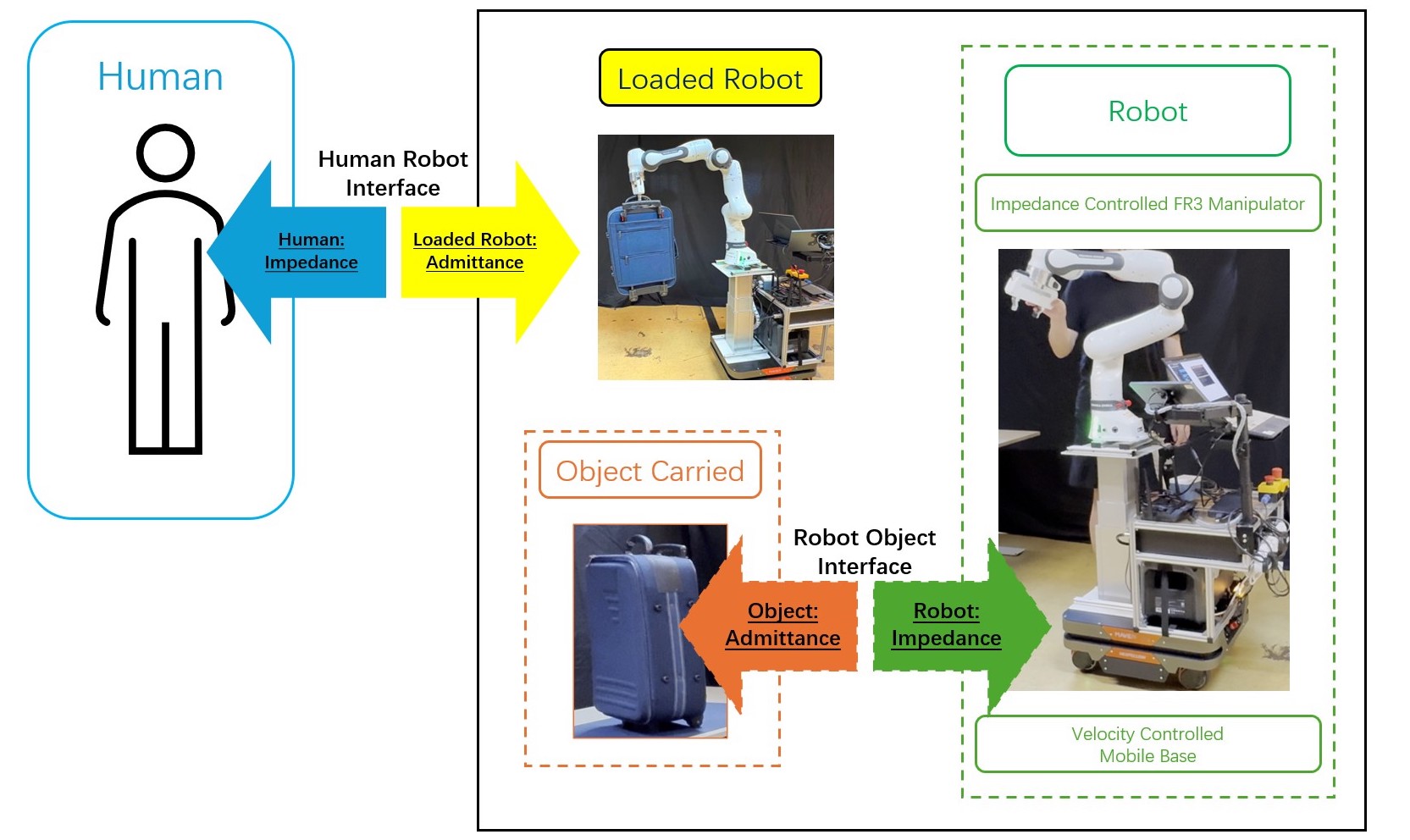}
    \caption{Physical roles during human-guided load transfer using a mobile manipulator.}
    \label{fig:physical_roles}
\end{figure}

The proposed controller architecture for the mobile manipulator is motivated by the physical roles involved in human-guided load transfer. As illustrated in Fig.~\ref{fig:physical_roles}, the robot interacts with both the human and the carried object. At the human-robot interface, the loaded robot should behave as an admittance: the human applies force, and the loaded robot system moves in response. At the robot-object interface, the manipulator should behave as an impedance: it supports the load compliantly around a commanded equilibrium pose.

\begin{figure}[!htbp]
\centering
\includegraphics[width=\textwidth]{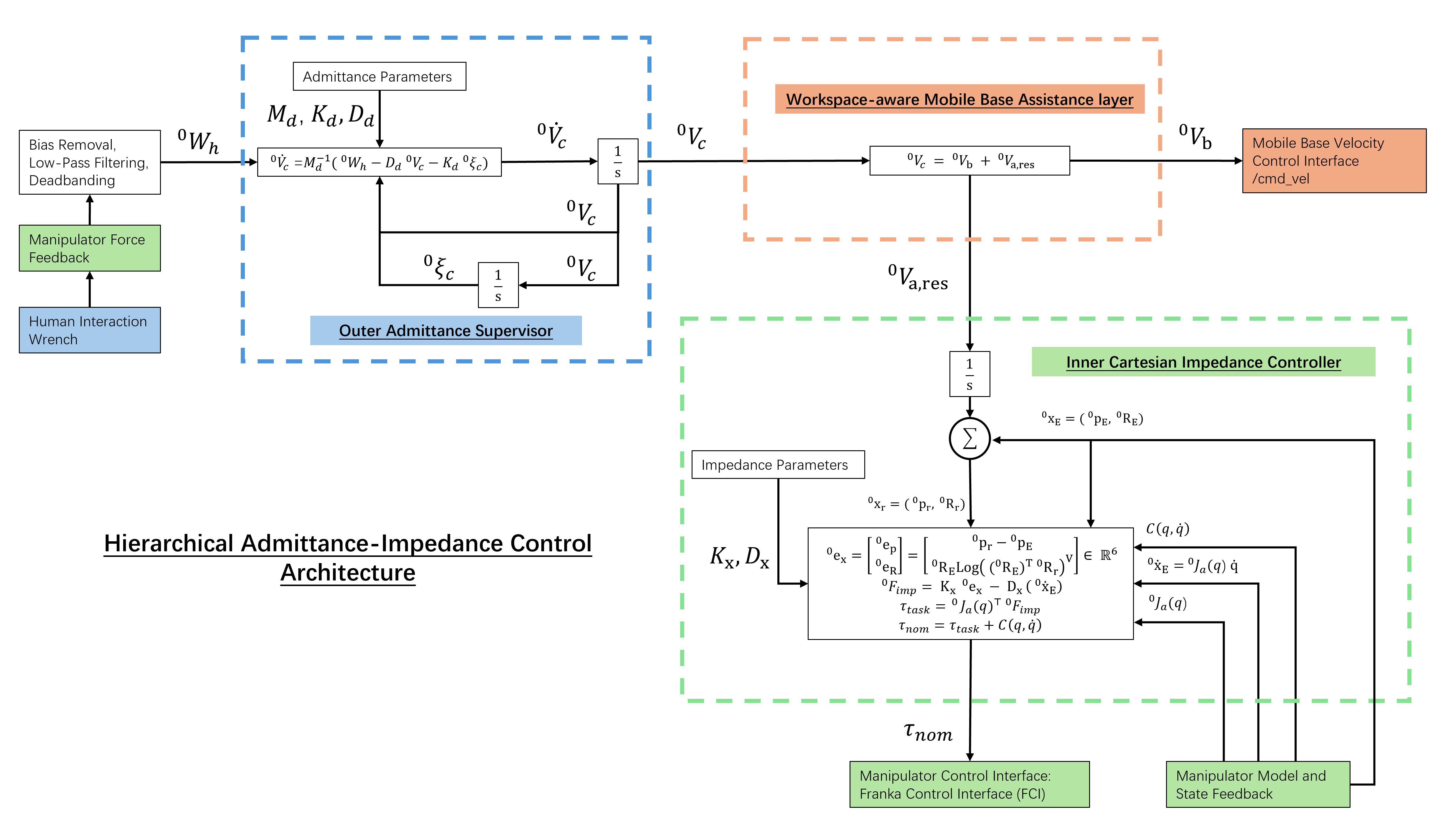}

\caption{Hierarchical admittance-impedance control architecture for human-guided load transfer with workspace-aware mobile-base assistance.}
\label{fig:control_architecture}
\end{figure}

This interaction-role interpretation leads to a three-layer controller, as illustrated in Fig.~\ref{fig:control_architecture}
\begin{enumerate}
    \item \textbf{Inner Cartesian impedance controller}, where the torque-controlled FR3 manipulator tracks the moving Cartesian equilibrium pose while maintaining local compliant behavior
    \item \textbf{Outer 6D admittance supervisor}, which converts the measured human interaction wrench into smooth compliant end-effector motion
    \item \textbf{Workspace-aware mobile-base assistance layer}, which shares the translational part of this motion between the manipulator reference and the mobile base using workspace activation, motion intent, and frontal load alignment.
\end{enumerate}

With this hierarchy, the human-applied wrench does not directly command joint torques or mobile base velocities. Instead, it moves the Cartesian equilibrium of the impedance-controlled manipulator through the admittance supervisor, while the mobile base assists when larger workspace motion or frontal load alignment is required.

\subsection{Inner Cartesian Impedance Controller}

The inner controller receives a Cartesian reference
\begin{equation}
{}^{0}\mathbf{x}_{r}
=   
\left(
{}^{0}\mathbf{p}_{r},
{}^{0}\mathbf{R}_{r}
\right)
\end{equation}
where ${}^{0}\mathbf{p}_{r}\in\mathbb{R}^{3}$ is the reference position and ${}^{0}\mathbf{R}_{r}\in SO(3)$ is the reference orientation. The measured end-effector pose is
\begin{equation}
{}^{0}\mathbf{x}_{E}
=
\left(
{}^{0}\mathbf{p}_{E},   
{}^{0}\mathbf{R}_{E}
\right)
\end{equation}

The Cartesian pose error is defined as
\begin{equation}
{}^{0}\mathbf{e}_{x}
=
\begin{bmatrix}
{}^{0}\mathbf{e}_{p} \\
{}^{0}\mathbf{e}_{R}
\end{bmatrix}
\in \mathbb{R}^{6}
\end{equation}
where ${}^{0}\mathbf{e}_{p}$ is the position error 
(${}^{0}\mathbf{e}_{p}
=
{}^{0}\mathbf{p}_{r}
-
{}^{0}\mathbf{p}_{E}$) and ${}^{0}\mathbf{e}_{R}$ is the orientation error ($
{}^{0}\mathbf{e}_{R}
=
{}^{0}\mathbf{R}_{E}
\mathrm{Log}
\left(
({}^{0}\mathbf{R}_{E})^{\top}
{}^{0}\mathbf{R}_{r}
\right)^{\vee}$).

The desired Cartesian impedance wrench expressed in $\mathcal{F}_{0}$ is

\begin{equation}
{}^{0}\mathbf{F}_{imp}
=
\mathbf{K}_{x}
{}^{0}\mathbf{e}_{x}
-
\mathbf{D}_{x}
\left(
{}^{0}\dot{\mathbf{x}}_{E}
\right)
\end{equation}

where $\mathbf{K}_{x}\in\mathbb{R}^{6\times 6}$ and $\mathbf{D}_{x}\in\mathbb{R}^{6\times 6}$ are the Cartesian stiffness and damping matrices. ${}^{0}\dot{\mathbf{x}}_{E}\in \mathbb{R}^{6}$ is the instantaneous end effector velocity calculated by ${}^{0}\mathbf{J}_{a}(\mathbf{q}) \dot{\mathbf{q}}$.

The task-space wrench is mapped into joint torque through the manipulator Jacobian:
\begin{equation}
\boldsymbol{\tau}_{task}
=
\left(
{}^{0}\mathbf{J}_{a}(\mathbf{q})
\right)^{\top}
{}^{0}\mathbf{F}_{imp}
\end{equation}
where ${}^{0}\mathbf{J}_{a}(\mathbf{q})\in\mathbb{R}^{6\times 7}$ is the FR3 geometric Jacobian expressed in $\mathcal{F}_{0}$.

The nominal torque command is
\begin{equation}
\boldsymbol{\tau}_{nom}
=
\boldsymbol{\tau}_{task}
+
\mathbf{C}
\left(
\mathbf{q},
\dot{\mathbf{q}}
\right)
\end{equation}
where $\mathbf{C}(\mathbf{q},\dot{\mathbf{q}})$ is the Coriolis compensation term (computed by the Franka Control Interface (FCI)). This layer gives the manipulator local impedance behavior while allowing the outer supervisor to move the equilibrium pose.

\subsection{Outer 6D Admittance Supervisor}

Let the measured human interaction wrench expressed in $\mathcal{F}_{0}$ be
\begin{equation}
{}^{0}\mathbf{W}_{h}
=
\begin{bmatrix}
{}^{0}\mathbf{f}_{h} \\
{}^{0}\boldsymbol{\tau}_{h}
\end{bmatrix}
\in \mathbb{R}^{6},
\end{equation}
where ${}^{0}\mathbf{f}_{h}\in\mathbb{R}^{3}$ is the interaction force and ${}^{0}\boldsymbol{\tau}_{h}\in\mathbb{R}^{3}$ is the interaction torque.

Let ${}^{0}\mathbf{V}_{c}$ be the desired compliant end-effector twist
\begin{equation}
{}^{0}\mathbf{V}_{c}
=
\begin{bmatrix}
{}^{0}\mathbf{v}_{c} \\
{}^{0}\boldsymbol{\omega}_{c}
\end{bmatrix}
\in \mathbb{R}^{6}.
\end{equation}


After bias removal, low-pass filtering, and deadbanding, the wrench drives a 6D admittance model:
\begin{equation}
\mathbf{M}_{d}
{}^{0}\dot{\mathbf{V}}_{c}
+
\mathbf{D}_{d}
{}^{0}\mathbf{V}_{c}
+
\mathbf{K}_{d}
{}^{0}\boldsymbol{\xi}_{c}
=
{}^{0}\mathbf{W}_{h},
\end{equation}
where $\mathbf{M}_{d}$, $\mathbf{D}_{d}$, and $\mathbf{K}_{d}$ are the virtual admittance inertia, damping, and stiffness matrices. 
The compliant displacement state satisfies ${}^{0}\dot{\boldsymbol{\xi}}_{c}= {}^{0}\mathbf{V}_{c}$.

For hand-guiding behavior, the stiffness term may be set to zero or chosen to be weak:
\begin{equation}
\mathbf{M}_{d}
{}^{0}\dot{\mathbf{V}}_{c}
+
\mathbf{D}_{d}
{}^{0}\mathbf{V}_{c}
=
{}^{0}\mathbf{W}_{h}.
\end{equation}

In discrete time, the admittance update is
\begin{equation}
{}^{0}\dot{\mathbf{V}}_{c}^{k}
=
\mathbf{M}_{d}^{-1}
\left(
{}^{0}\mathbf{W}_{h}^{k}
-
\mathbf{D}_{d}
{}^{0}\mathbf{V}_{c}^{k}
-
\mathbf{K}_{d}
{}^{0}\boldsymbol{\xi}_{c}^{k}
\right),
\end{equation}
\begin{equation}
{}^{0}\mathbf{V}_{c}^{k+1}
=
{}^{0}\mathbf{V}_{c}^{k}
+
\Delta t
{}^{0}\dot{\mathbf{V}}_{c}^{k},
\end{equation}
\begin{equation}
{}^{0}\boldsymbol{\xi}_{c}^{k+1}
=
{}^{0}\boldsymbol{\xi}_{c}^{k}
+
\Delta t
{}^{0}\mathbf{V}_{c}^{k+1}.
\end{equation}

The output ${}^{0}\mathbf{V}_{c}$ is the desired compliant twist of the robot-object system, which is not directly assigned to the manipulator. 
In compact form,
\begin{equation}
{}^{0}\mathbf{V}_{c}
=
{}^{0}\mathbf{V}_{b}
+
{}^{0}\mathbf{V}_{a,\mathrm{res}}
\end{equation}
where ${}^{0}\mathbf{V}_{b}$ is the end-effector motion contribution associated with the mobile base motion, and ${}^{0}\mathbf{V}_{a,\mathrm{res}}$ is the residual twist used to update the Cartesian equilibrium pose of the impedance-controlled manipulator.

\subsection{Workspace-Aware Mobile Base Assistance Layer}



The workspace-aware mobile-base assistance layer extends the effective workspace of the robot while maintaining a favorable carrying configuration. The outer admittance supervisor generates ${}^{0}\mathbf{V}_{c}$, and the assistance layer assigns part of the horizontal translational motion (${}^{0}\mathbf{V}_{b}$) to the mobile base when the load approaches predefined manipulator workspace limits. The remaining motion (${}^{0}\mathbf{V}_{a,\mathrm{res}}$) is used to control the manipulator. 

In addition, frontal load alignment commands the mobile base's yaw motion when the payload drifts laterally, helping keep the payload in front of the robot for safer and more predictable load transfer.

\subsection{Operational sequence}

\begin{figure}[!htbp]
    \centering
    \includegraphics[width=\linewidth]{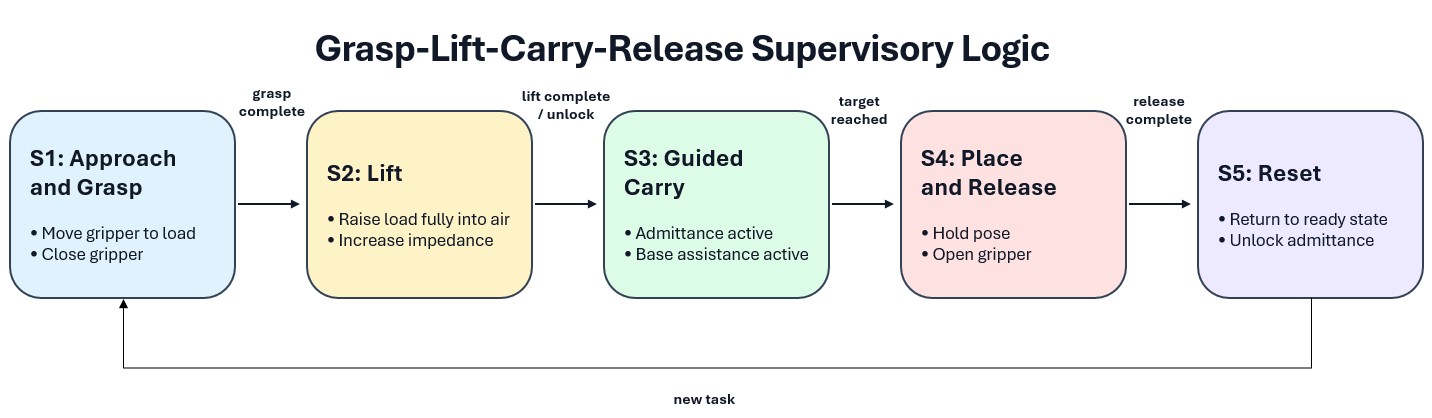}
    \caption{Supervisory logic for robot-assisted load transfer}
    \label{fig:task_logic}
\end{figure}

The supervisor also includes a simple task-state logic for load transfer, as shown in Fig.~\ref{fig:task_logic}. The sequence consists of five states: approach and grasp, lift, guided carry, place and release, and reset.

In the approach and grasp state, the gripper is moved to the payload and closed. In the lift state, the admittance motion is temporarily locked, the reference is moved upward until the load is fully in the air, and the controller impedance is increased for carrying. After lifting, the payload baseline is refreshed, and admittance is unlocked. During guided carry, the human guides the suitcase while the robot supports the load, and the mobile base assistance remains active. In the place and release state, the robot holds the pose and opens the gripper. Finally, the reset state returns the system to a ready condition and unlocks admittance for the next task.

This sequence provides a repeatable procedure for grasping, lifting, carrying, and releasing the load while the continuous admittance-impedance controller handles physical guidance.

\section{Experimental Evaluation}

\subsection{Objective}

The experiment evaluates the proposed hierarchical admittance-impedance controller in a load transfer task with robot assistance. The objective is to demonstrate that the mobile manipulator can support and transfer a suitcase while a human operator guides the motion through physical interaction. 
    
\subsection{Experimental Setup and Procedure}

\begin{figure}[t]
    \centering
    \includegraphics[width=\linewidth]{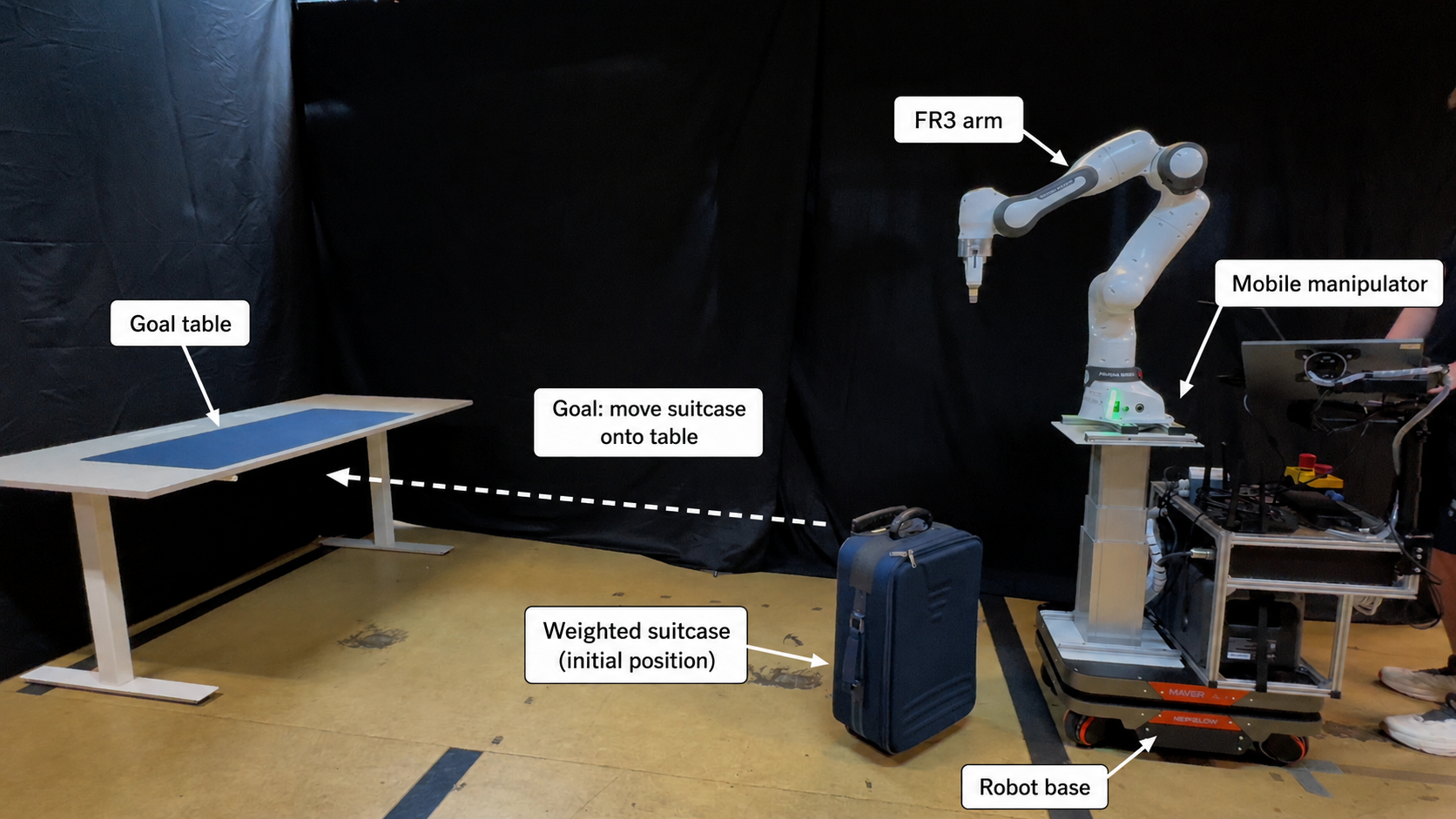}
    \caption{Experimental Setup}
    \label{fig:experiment_setup}
\end{figure}

The experimental setup is shown in Fig.~\ref{fig:experiment_setup}, where a Franka Research 3 manipulator is mounted on a mobile base. The manipulated object is a suitcase with a measured mass of approximately $3.3~\mathrm{kg}$, chosen as a safe and repeatable object for validating the proposed human-guided load-transfer behavior under conservative payload limits. The task goal is to move the suitcase from its initial position onto a target table, where the human operator physically guides the suitcase through direct interaction while the robot supports the load and moves compliantly according to the proposed admittance-impedance controller.

The experiment consists of five stages, with the same controller used throughout while the supervisor switches between task states as illustrated in Fig.~\ref{fig:task_logic}.

\begin{enumerate}
    \item \textbf{Stage 1: Move gripper to suitcase:} the robot manipulator is guided from its initial configuration toward the suitcase handle or grasping region.
    \item \textbf{Stage 2: Lift load:} the robot grasps the suitcase and lifts it by a prescribed height.
    \item \textbf{Stage 3: Human-guided load transfer:} the human operator guides the loaded robot system toward the target table through physical interaction.
    \item \textbf{Stage 4: Release load:} the suitcase is placed on the target table and released.
    \item \textbf{Stage 5: Unloaded return:} after release, the unloaded robot is moved back to its initial configuration.
\end{enumerate}

Throughout the experiment, parameters such as elapsed time, total end-effector force magnitude, human guidance force magnitude, estimated stable load weight, manipulator extension length, cumulative global load travel, and load angle relative to the robot's frontal direction were recorded to facilitate performance evaluation of the proposed controller.



\subsection{Evaluation Metrics}

The quantitative evaluation focuses on Stage~3, from $24~\mathrm{s}$ to $52~\mathrm{s}$, where the suitcase has already been lifted and the human physically guides the loaded robot-suitcase system toward the table. The following metrics are used to evaluate controller performance.

\begin{itemize}
    \item \textbf{Human guidance force:} The primary indicator of human effort during the transfer. A guidance force significantly smaller than the suitcase weight indicates that the robot supports the load while the human mainly provides directional input.
    \item \textbf{Manipulator extension:} The reach of the manipulator during the task. A bounded extension verifies that the manipulator neither over-stretches near its workspace limit nor collapses too close to the mobile base.
    \item \textbf{Global load travel distance:} The total distance the suitcase is transported during the guided transfer, confirming that the system performs a meaningful load-transfer task beyond local manipulator motion.
    \item \textbf{Absolute load angle:} The load angle relative to the robot's frontal direction. Keeping this angle small ensures the suitcase remains in front of the robot, which is important for safety as it keeps the center of gravity closer to the robot's forward support region and reduces lateral side loading of the manipulator.
\end{itemize}

\section{Results}

\begin{figure}[!htbp]
    \centering
    \includegraphics[width=\linewidth,height=1.0\textheight,keepaspectratio]{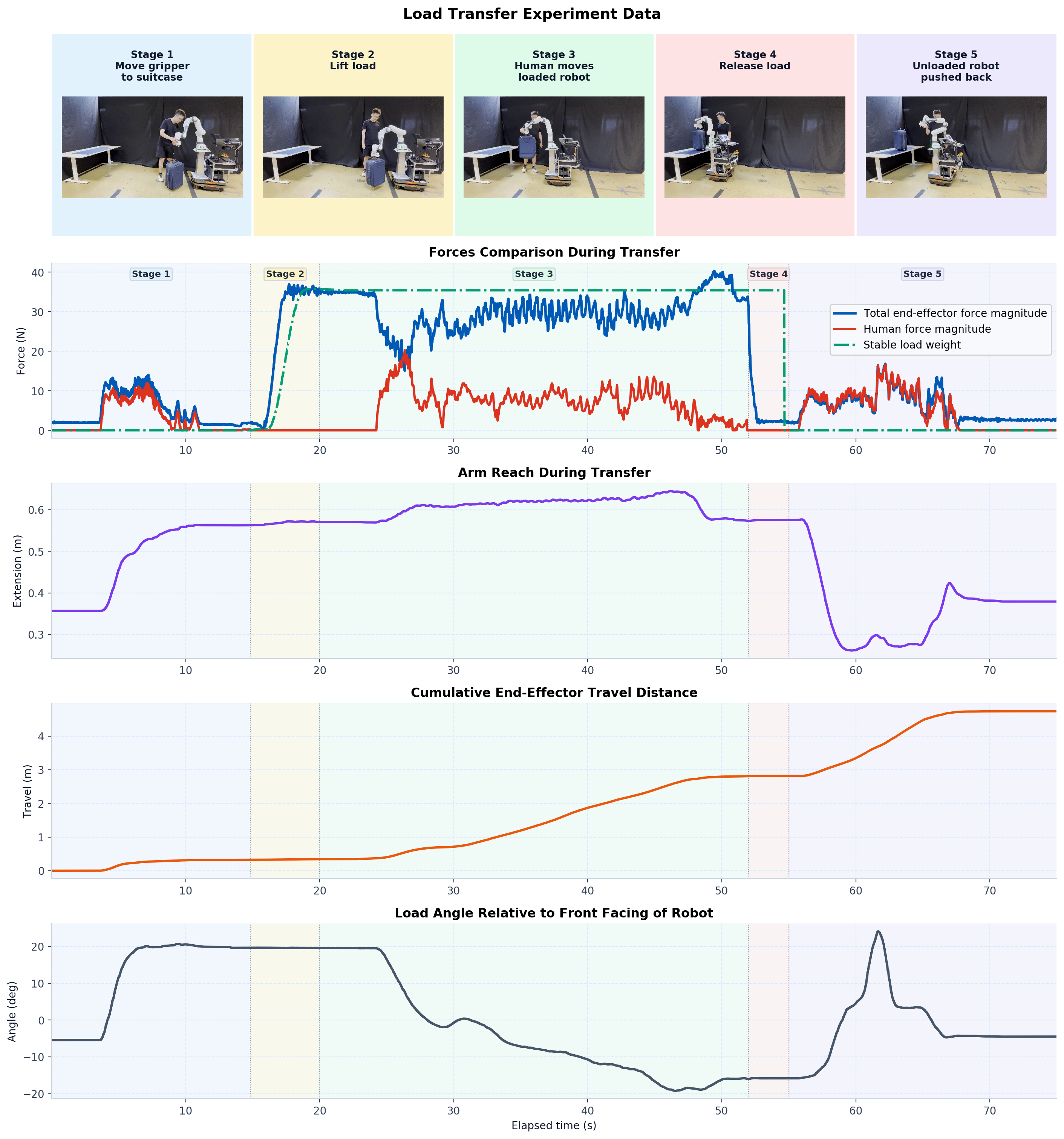}
    \caption{Recorded data from the robot-assisted suitcase transfer experiment.}
    \label{fig:experiment_results}
\end{figure}

Figure~\ref{fig:experiment_results} shows the recorded data from the robot-assisted suitcase transfer experiment, and Table~\ref{tab:stage3_results} summarizes the quantitative indicators during Stage~3, reporting the average, minimum, and maximum values over the evaluation interval.

\begin{table}[t]
\centering
\caption{Quantitative results during Stage~3 human-guided loaded transfer from $24~\mathrm{s}$ to $52~\mathrm{s}$.}
\label{tab:stage3_results}
\begin{tabular}{lccc}
\hline
Indicator & Average & Minimum & Maximum \\
\hline
Human guidance force (N) & 7.43 & 0.00 & 20.22 \\
Total end-effector force (N) & 29.49 & 15.11 & 40.31 \\
Manipulator extension (m) & 0.612 & 0.570 & 0.645 \\
Absolute load angle (deg) & 10.31 & 0.01 & 19.54 \\
\hline
\end{tabular}
\end{table}

The suitcase mass was approximately $3.3~\mathrm{kg}$, corresponding to a gravitational weight of about $32.4~\mathrm{N}$, while the estimated stable load weight during Stage~3 was approximately $35.40~\mathrm{N}$. The average human guidance force was $7.43~\mathrm{N}$, approximately $21.0\%$ of the supported load, confirming that the robot carried the main payload while the human provided directional guidance. The peak guidance force of $20.22~\mathrm{N}$ occurred during a transient motion where the operator simultaneously drove the system forward and adjusted the end effector upward, and therefore reflects a short corrective command rather than the steady effort required to support the suitcase weight. Over the course of Stage~3, the cumulative load travel increased from $0.366~\mathrm{m}$ to $2.809~\mathrm{m}$, corresponding to a total transfer distance of approximately $2.44~\mathrm{m}$.

The manipulator extension remained within a bounded range throughout Stage~3, with an average of $0.612~\mathrm{m}$ and a range of $0.570$--$0.645~\mathrm{m}$, showing that the arm neither over stretched closer to its workspace boundary nor collapsed too close to the mobile base. The average absolute load angle was $10.31^{\circ}$, with a maximum of $19.54^{\circ}$, indicating that the suitcase remained generally in front of the robot and within a safe lateral alignment throughout the transfer.

Together, these results demonstrate that the proposed hierarchical admittance-impedance controller successfully facilitates human-guided load transfer: the robot supported approximately $35.40~\mathrm{N}$ of payload while the human applied an average guidance force of only $7.43~\mathrm{N}$, the suitcase travelled $2.44~\mathrm{m}$, and both the manipulator extension and load angle remained within safe bounds throughout the task.

\section{Discussion}

The proposed controller provides a practical way to combine human guidance, compliant manipulation, and mobile base assistance in a single system. One design advantage is the separation between the inner impedance-controlled manipulator and the outer admittance supervisor. The manipulator remains locally compliant and torque controlled, while the human interaction force is interpreted at the supervisory level as a motion command. This makes the system intuitive to guide without requiring the human to command robot joints or specify a predefined trajectory.

Another advantage is that the method can be implemented on a mobile manipulator with heterogeneous control interfaces. The FR3 manipulator is controlled through joint torques, whereas the mobile base accepts velocity commands. Instead of requiring a unified whole-body torque controller, the proposed hierarchy connects these two subsystems through reference motion generation and workspace-aware base assistance. This makes the approach easier to deploy on various mobile manipulator platforms.

The workspace-aware base assistance also improves the usefulness of the system for load transfer tasks. The manipulator can provide compliant local support, while the mobile base contributes to large scale motion. This allows the robot to move the load over a larger distance without relying only on manipulator extension. In addition, the frontal load alignment behavior keeps the payload in a more predictable carrying region, which is important for safety and operator confidence during physical guidance.

The current implementation is intentionally simple and modular. The admittance parameters, impedance parameters, workspace thresholds, base velocity limits, and load alignment gains can be tuned independently. This makes the controller easy to test and adapt, but it also means that performance depends on parameter selection. Future work should therefore investigate more systematic tuning of admittance and impedance parameters. This could include task dependent parameter scheduling, optimization based tuning, or learning based adaptation from human demonstration and interaction data.

Another important direction is improving autonomy in the grasping stage. In the current experiment, the robot is guided to the suitcase before the load transfer phase. Future work should integrate perception and grasp planning so that the robot can autonomously detect the load, approach it, and grasp it without requiring the human to guide the gripper to the object. This would make the system more complete for practical assistive load transfer applications.

\section{Conclusion}

This paper presented a hierarchical admittance-impedance controller for human guided load transfer with a mobile manipulator. The method combines an inner Cartesian impedance controller for a torque-controlled Franka Research 3 manipulator, an outer admittance supervisor for force guided reference generation, and a workspace-aware mobile base assistance strategy with frontal load alignment.

The controller was evaluated in a robot assisted suitcase transfer experiment. The results showed that the robot can support the payload while the human guides the motion with relatively small physical effort. The experiment also showed that the mobile base can assist the load transfer over a larger distance while keeping the manipulator within a bounded reach range and maintaining the load generally in front of the robot.

Overall, the proposed framework provides a simple and implementable approach for compliant human guided mobile manipulation. It preserves direct human authority over the object motion while allowing the robot to reduce the physical burden of carrying and transferring loads. Future work will focus on systematic admittance and impedance parameter tuning, learning based adaptation, and autonomous grasping to further improve the practicality of the system for assistive load transfer tasks.

%
%

\end{document}